\documentclass[11pt]{article}
\usepackage[preprint]{acl}

\usepackage{times}
\usepackage{latexsym}

\usepackage[T1]{fontenc}
\usepackage[utf8]{inputenc}
\usepackage{microtype}
\usepackage{inconsolata}
\usepackage{graphicx}

\usepackage{amsmath}
\usepackage{amsfonts}
\usepackage{amssymb}
\usepackage{amsthm}
\usepackage{subcaption}
\usepackage[title]{appendix}

\title{Comment-level Topic Drift Analysis in the Reddit Corpus}

\author{Steven Morse \and Dan Runfola \and Trenton W. Ford \\
  William \& Mary  \\
  \texttt{\{stmorse, dsmillerrunfol, twford\}}@wm.edu}

\begin{document}

\maketitle

\begin{abstract}
We present a novel application of embedding-based dynamic topic modeling techniques to detect and quantify topic drift at the comment level in a massive corpus.  By leveraging pretrained language models to generate contextualized semantic embeddings for short text, we analyzed 12.7 billion Reddit comments spanning 2006 to 2022. Using unsupervised methods on these embeddings, we identify dynamically evolving topic clusters over time.  Our primary contribution is a methodology for analysis of semantic drift and discourse evolution in the embedding space itself.  We also demonstrate modifications to existing methods that enable this analysis at scale, and we propose and demonstrate a null model comparison test to filter spurious dynamics. Key findings suggest that politically and socially contentious topics exhibit significant directional drift in embedding space, with inter-topic distances changing systematically over time beyond what the null model can explain, whereas domains such as music and sports remain comparatively stable.
\end{abstract}


\section{Introduction}\label{intro}

Attention-based language models encode language into a high-dimensional embedding space that captures both semantic relationships and long-range contextual dependencies, enabling a nuanced representation of meaning. This has enabled the study of topic analysis as computation in a vector space \cite{churchill2022evolution,wu2024survey}. Beyond static clustering, embedding-based methods enable quantitative comparisons among words, sentences, and topics \cite{grootendorst2022bertopic}. Yet most pipelines remain descriptive: they list topics or track keyword changes, but they do not estimate whether topics move meaningfully through embedding space over time or compare those movements across topics. That gap matters because the contextualized semantics of topic discourse shifts with events and norms; without a time-resolved geometric model, we cannot measure or forecast that change.

We address this gap by treating change in topical discourse as trajectories in embedding space and asking when observed movement exceeds stochastic variation. Concretely, we embed comments, cluster within monthly windows, align similar clusters across time to recover topic paths, and test those paths against a null model to separate genuine drift from noise. At web-scale over multiple decades, this yields interpretable measures of displacement and direction and exposes inter-topic convergence and divergence --- for example, two topics may retain the same representative words (for example, ``police'' and ``racism'') but grow closer in semantic space, implying a change in semantic similarity of content not detectable by methods treating topics as bags of words. The result is a practical framework for tracking discourse change, evaluating interventions, and informing downstream tasks such as retrieval, moderation, and forecasting.

\begin{figure*}[t]
    \centering
    \includegraphics[width=0.75\textwidth]{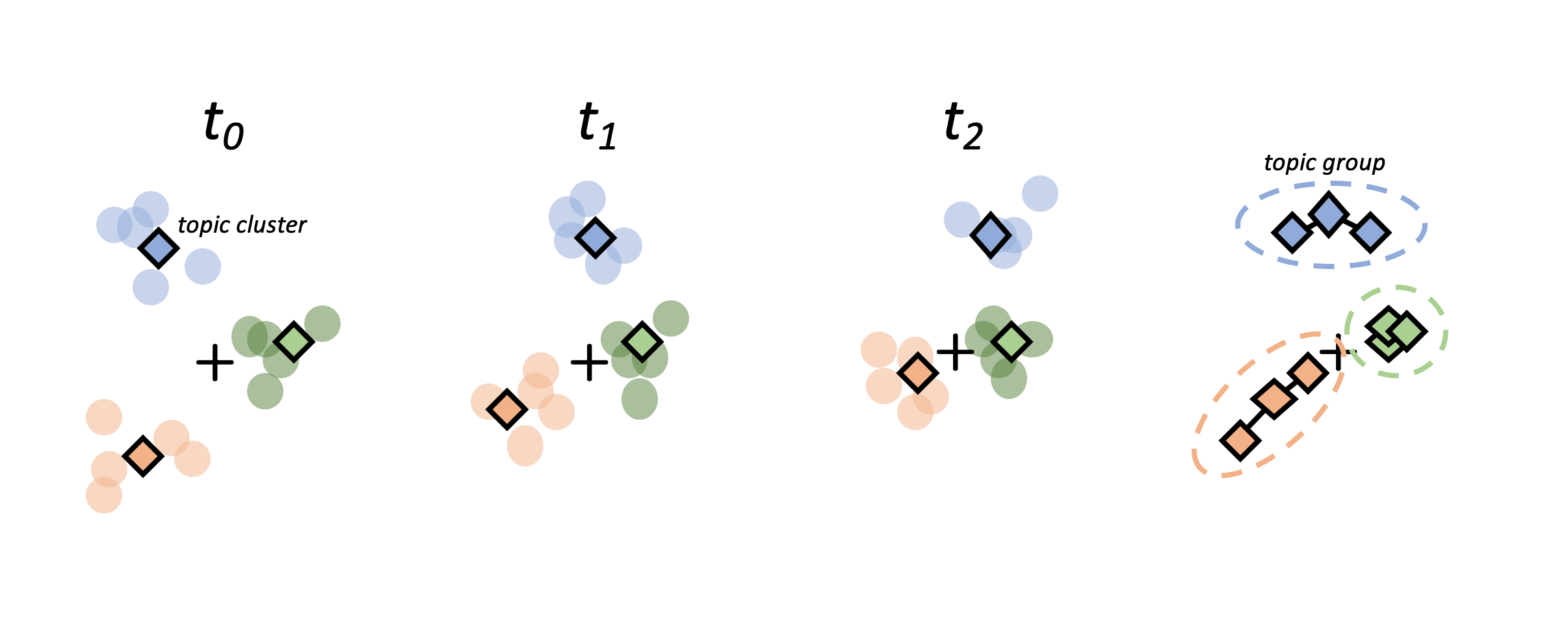}
    \caption{Clustering in embedded space yields time-local topics with contextualized semantic similarity over time.  Aligning highly similar topic clusters over time may reveal {\it topic drift} in embedding space.}
\end{figure*}


\subsection{Related work}

Topic modeling identifies and organizes latent themes in a collection of texts \cite{vayansky2020review}, and it categorizes broadly into approaches based on distributional language frameworks and those leveraging deep neural network embeddings of language \cite{churchill2022evolution}. Within the distributional paradigm, considerable research has focused on modeling temporal dynamics and, as a natural extension, analyzing changes in topic distributions (``drift'') \cite{khan2021drifting,knights2009detecting,liu2013micro}.  Within the neural, or algorithmic paradigm, similar extensions exist, with most studies focused at the word (or token) level \cite{bamler2017dynamic,periti2024lexical}.  In this section, we review this literature and introduce our own contribution: the study of drift at the sentence (or in our context, comment) level, in an embedded space and at large scale.

Probabilistic approaches typically approach topics as a collection of independently occurring words (``bag of words'') \cite{churchill2022evolution}.  Latent Dirichlet Allocation (LDA) \cite{blei2003latent} introduced a flexible, hierarchical Bayesian framework for this task, with subsequent work exploring metrics of topic coherence and diversity \cite{roder2015exploring}, expansion to a dynamic setting where topics' distributions change over time \cite{blei2006dynamic,wang2012continuous,bhadury2016scaling,zhang2022dynamic,iwata2010online}, and analysis of distributional drift in the dynamic setting \cite{knights2009detecting,khan2021drifting,liu2013micro,li2017learning,fei2015handling}.  In work extending these topic models to larger datasets, we find modifications to existing methods \cite{yuan2015lightlda,hassani2020text}, applications to marketing \cite{amado2018research}, social media \cite{guo2016big}, among many others.  We also see the application of modern neural network architectures (``neural topic models'' (NTMs) \cite{wu2024survey}), which in most cases extends probabilistic approaches by using vectorized BoW input to a neural net, as in \cite{miao2016neural,srivastava2017autoencoding,krishnan2018challenges,zhao2020neural}.

More relevant to our work is the recent field of approaches that leverage transformer-based language models' embeddings to capture deep contextual information about the text.  Early works like word2vec \cite{mikolov2013efficient} and GloVe \cite{pennington2014glove} found word embeddings using shallow neural networks trained to predict the immediate context of a given word.  Large Language Models (LLMs) based on the transformer architecture \cite{vaswani2017attention,kenton2019bert,zhou2024comprehensive}, provide even richer language embeddings.  Research has since extended this transformer approach to sentence-level embeddings \cite{incitti2023beyond} --- for example, the Sentence-BERT approach in \cite{reimers2019sentence2} involves fine-tuning a BERT model through contrastive training on a dataset of paired sentences, driving the embedding space's geometry to capture semantic similarity. This concept is highly adaptable, as evidenced by a growing suite of tools \cite{thielmann2024stream}.

These language embeddings enable new approaches in topic modeling by working in this embedding space \cite{wu2024survey}.  For example, we may apply unsupervised methods directly to identify topic clusters, either at word \cite{xie2019method} or sentence level \cite{grootendorst2022bertopic}, which we discuss further below.  We may extend this to analysis of word dynamics \cite{bamler2017dynamic,yao2018dynamic,hofmann2020dynamic} and semantic shift detection \cite{periti2024lexical,taher2023embedding}.  Many NTM approaches also work by incorporating embeddings into a BoW model (e.g., \cite{bunk2018welda}), either as a kind of ``side-information'' \cite{petterson2010word} or directly as input \cite{dieng2019dynamic, dieng2020topic}, although \cite{zhang2022neural} argues that unsupervised approaches directly in the embedding space yield better results.  Note also there is a growing field of study around feature drift in LLMs (e.g. \cite{fastowski2024understanding,santos2024adaptive}), but this differs from our subject in its focus on model reliability and robustness in the context of feature drift, not topic modeling and semantic drift.


\subsection{Contributions}

We build on two embedding-based approaches to topic dynamics within this literature. BERTopic discovers time-agnostic clusters via global clustering, then inspects time-sliced keyword changes, which imposes a global view that can mask drift of topics in embedding space \cite{grootendorst2022bertopic}. The complementary method in \cite{rahimi2024antm} partitions time first, clusters locally, and aligns clusters across periods using hierarchical procedures. Both methods emphasize keyword or coherence summaries rather than geometry, provide no formal test that observed dynamics exceed stochastic variation, and rely on global dimensionality reduction that is impractical at web-scale.

Our work augments this literature in three ways.  First, we propose a more scalable comment-level pipeline: we perform topic clustering directly in the original embedding space within monthly windows, then perform dimensionality reduction and alignment of the resulting  centroids to discover topic trajectories.  Second, we introduce a likelihood-ratio test based on a random walk null model to filter genuine topic drift from noise.  This framework and null together provide quantitative measures of displacement and directionality both within and between topics, enabling interpretation in the embedding space.  Our third contribution is to demonstrate this analysis on a web-scale dataset, the Reddit corpus.

\begin{figure*}[htb]
    \centering
    \begin{subfigure}{0.49\textwidth}
        \includegraphics[width=\linewidth]{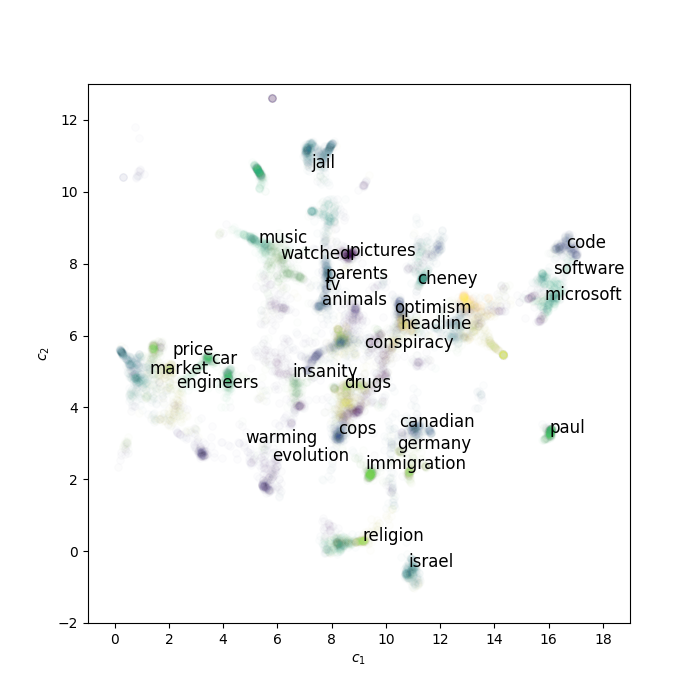}
        \caption{Topic clusters (June 2007)}
    \end{subfigure}
    \begin{subfigure}{0.49\textwidth}
        \includegraphics[width=\linewidth]{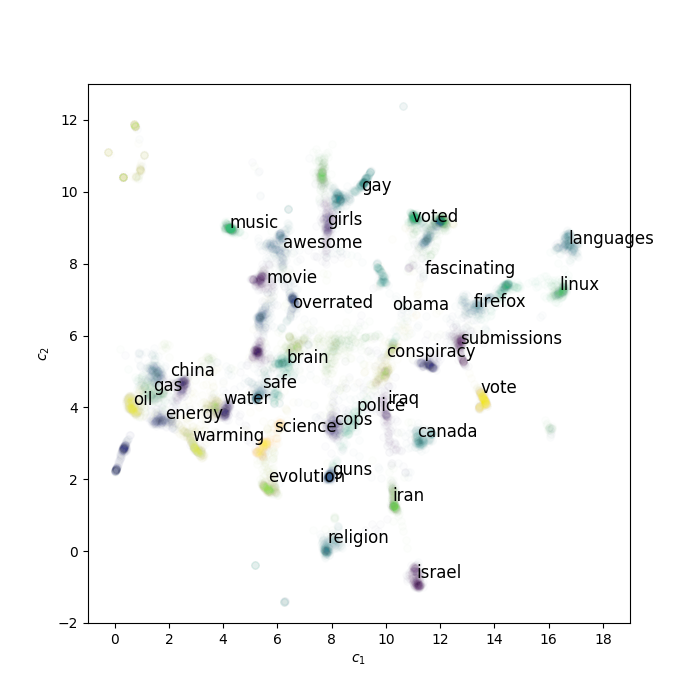}
        \caption{Topic clusters (June 2008)}
    \end{subfigure}
    \caption{Visualization of topic clusters for two example time periods (June 2007 {\bf (a)} and 2008 {\bf (b)}), with each dot representing a single embedded comment.  Only the top 200 closest to the centroid are shown for visual clarity; text label is the top scoring word by TF-IDF. We note trends of changing focus: increased presidential voting discussion, diversification of several topics (e.g., immigration), and increase in sexually-oriented commenting.}
    \label{fig:singlemonth}
\end{figure*}

\section{Data \& Methodology}\label{methods}

\subsection{Data}

We explore topic drift using embeddings of the Reddit comment corpus (2006–2022) released via Pushshift \cite{baumgartner2020pushshift}. To analyze topical discourse in text, we restrict to comments and exclude top-level submissions and multimedia. Comments authored by deleted accounts are removed. Very short replies (including single-word comments) are retained to preserve conversational structure; each comment is truncated to at most 128 tokens to match the embedding model’s context window.

After filtering, the corpus comprises $T=204$ monthly windows and 12.7\,billion comments. The volume is highly skewed toward recent years: the most recent five-year period contains $>$9\,billion comments, more than $160\times$ the first five years ($<60$\,million).\footnote{The corpus includes extremist, violent, and offensive content. Results are presented without reproducing disturbing material, but such material was included in the analysis.}

\subsection{Methodological framework}

This work leverages a framework consisting of three stages.  

\begin{enumerate}
    \item {\bf Contextualized embedding.} The first stage uses a pretrained, transformer-based LLM to convert a short body of text into a high-dimensional vector representation capturing contextual encoding.  We perform this embedding using a fixed transformer across the entire dataset.  
    
    \item {\bf Topic clustering.}  The second stage partitions the embeddings into time-windows and performs clustering within each window to infer topic clusters by grouping texts that are close in the embedding space.  Concurrent with this layer, we produce keyword representations of each topic cluster to aid in semantic analysis.
    
    \item {\bf Topic alignment.}  The third stage works across time windows to align similar topic clusters over time, in a locally-time-aware way.  Given these aligned topic groups, we analyze the resulting ``trajectories'' of topics in embedding space, specifically, their absolute drift and drift relative to each other.  We also conduct a dimensionality reduction prior to this alignment, and we note this differs from \cite{grootendorst2022bertopic,rahimi2024antm} which apply a global dimensionality reduction (UMAP) prior to clustering, a prohibitive step for massive datasets.
\end{enumerate}

We then perform analysis of the resulting aligned topic groups in the embedding space, leveraging the interpretation of distance in the embeddings space as a representation of semantic similarity.

\subsection{Contextual Embedding}

Consider a dataset $\mathcal{D}=\{(s_i, \hat{t}_i)\}$ consisting of ``mini-documents'' (generally of sentence to paragraph length) $s_i\in\mathcal{S}_\ell$ with timestamps $\hat{t}_i$ at high granularity (e.g., seconds), and where $\mathcal{S}_\ell$ represents character strings of maximum length $\ell$.  Our first step is to embed each mini-document in a space which captures semantic and contextual similarity, using a transformation $E: \mathcal{S}\rightarrow \mathcal{E}=\mathbb{R}^d$.  For $E$, we use a pretrained LLM following the SBERT and SentenceTransformer work in previous literature \cite{reimers2019sentence2}.  The resulting embeddings serve as vectorized representations of the sentences, optimized to enable accurate comparisons of semantic similarity. 

Specifically, we used `all-MiniLM-L6-v2`\footnote{Available at \href{https://huggingface.co/sentence-transformers/all-MiniLM-L6-v2}{huggingface.co}} which embeds each post $s_i$ into $\mathbb{R}^d$ with $d=384$ (following the methodology presented in \cite{rahimi2024antm}). The output of this step is an embedded dataset $\mathcal{D}_E = \{(\mathbf{e}_i, \hat{t}_i)\}$ consisting of time-stamped, embedded mini-documents $\mathbf{e}_i\in\mathcal{E}$, with $d$ representing the embedding dimension of the sentence transformer $E$.

\subsection{Time-windowed Topic Clustering}

After embedding the dataset, we partition it into discrete time windows $\mathcal{T}=\{0, 1, ..., T\}$ such that each time window $t$ consists of all embeddings $\mathcal{E}_t$ with timestamps $\hat{t}$ in the range $[t^{\text{min}}, t^{\text{max}})$.  We set the time window as a calendar month, and within each window, we assume there is a set of $M$ topics such that an embedding $\mathbf{e}_i$ is a sample from some topic $m$, which we denote as $\mathbf{e}_i^{(m)}$.  Following earlier studies \cite{gao2019jointly}, we further assume each topic has some underlying topic distribution for that time period, which is approximately normal in the embedding space, that is:
\begin{equation}
    \mathbf{e}_i^{(m)} \sim \mathcal{N}(\boldsymbol{\mu}_{t_i}^{(m)}, \mathbf{\Sigma}_{t_i}^{(m)}).
\end{equation}
with $\boldsymbol{\mu}_{t_i}^{(m)}\in\mathbb{R}^d$ and $\mathbf{\Sigma}_{t_i}^{(m)}\in\mathbb{R}^{d\times d}$.

To identify topic assignments in an unsupervised way, we perform a clustering step within each time window $C:\mathcal{E}_t\rightarrow \mathcal{L}_t$ which assigns a label $l_i$ to each embedding $\mathbf{e}_i$.  Any unsupervised method that yields a labeling of the inputs may be employed at this stage; however, under the normality assumption, a Gaussian mixture model is particularly well-suited. 

With this consideration in mind, we apply $k$-means clustering.  K-means is well-known to approximate a GMM with spherical covariance (e.g., \cite{mackay2003information}); in addition, it scales sufficiently to operate within the full $d$-dimensional embedding space and handle an extensive corpus. Because the embedding space encodes contextually informed semantic similarity, we may interpret the resulting clusters as topic clusters.  We find that $k=50$ clusters provide a consistent representation of topics across the entire dataset, and we provide a more thorough analysis of this choice in Appendix \ref{app:k}.

The output of this step is a labeled dataset $\mathcal{D}_L=\{(\mathbf{e}_i, t_i, l_i)\}$.  Note that our clustering method allows us to associate each label $l_i\in L=\{1, ..., M\}$ within a time window $t$ with a centroid $\mathbf{c}_{t}^{(m)}$ that estimates the topic mean, $\boldsymbol{\mu}_{t}^{(m)}$. Note also each specific timestamp $\hat{t}_i$ is replaced with the corresponding time window (i.e. month) $t_i$.

Since we perform this unsupervised clustering stage independently within each time window, we keep open the possibility of capturing different topic clusters in different windows, and avoid imposing a global view on the embedded location of any particular topic (cf. \cite{rahimi2024antm}).

To improve semantic interpretability, we assign a single-keyword summary to represent each cluster. This is implemented using the topic-frequency inverse-document-frequency (TF-IDF) approach, treating each cluster as a document, and normalizing frequency by time-window (c.f. \cite{grootendorst2022bertopic}).  Specifically, define
\begin{align}
\text{tf}(t,c) &= \frac{f_{t,c}}{\sum_{t'\in c} f_{t',c}} \\
\text{idf}(t,C) &= \log \frac{N}{|\{c:c\in C, \ t\in c\}|}
\end{align}
where $\text{tf}(t,c)$ captures how relevant term $t$ is to document $c$, and $\text{idf}(t,C)$ captures how unique term $t$ is across all documents $C$ by its ratio of occurrence $N$ in $c$ to all documents.  In our application, we treat each cluster as a document $c$, and all clusters in a time-window as $C$.  In practice, since we are primarily concerned with interpretability, we use only the $|c|=100$ mini-documents nearest the centroid $\mathbf{c}_i$ as the ``$c_i$'' for TF-IDF purposes.  We emphasize that these keywords are only to aid in human interpretability, as we perform all analysis and validation in the embedding space itself.

\subsection{Topic Alignment}

Because clustering is performed independently within each time period, a subsequent alignment stage is required to identify clusters that represent the same underlying topic \textit{across} periods. This approach contrasts with the global clustering of \cite{grootendorst2022bertopic} and follows the temporal alignment strategy of \cite{rahimi2024antm}. In essence, the goal is to determine whether topic $a$ at time $t$ and topic $b$ at time $t+1$, produced by separate period-specific clusterings, correspond to the same semantic concept.

To accomplish this, given the labeled dataset $\mathcal{D}_L=\{(\mathbf{e}_i, t_i, l_i)\}$, we perform an additional unsupervised step that clusters the topic centroids $\mathbf{c}_t^{(m)}$ to align similar topics over time. Because no invariant topic representation exists across periods, alignment is based on proximity in embedding space rather than on shared keywords or TF-IDF features. This approach preserves contextual semantics and enables detection of topic evolution, emergence, and disappearance over time. We specifically apply the Uniform Manifold Approximation and Projection (UMAP) algorithm to reduce the centroid embeddings to $d=10$ dimensions, followed by Hierarchical DBSCAN (HDBSCAN) clustering to group temporally related topics. This combination allows flexible detection of gradual semantic drift and heterogeneous topic structures, yielding a final mapping $A: \mathcal{C} \rightarrow \mathcal{G}$ from centroids to topic groups and a dataset $D_A = \{(\mathbf{c}_i, g_i)\}$ representing each topic and its aligned group across time.

\subsection{Validation of drift}

A central question to our work is assessing topics' drift through embedding space over time.  A limited but straightforward approach to measure this is to measure a topic group's total displacement over time.  That is, for group $k$ define its displacement $\delta_k$ as
\begin{equation}
    \delta_k \stackrel{\text{def}}{=} ||\mathbf{c}_{t_{\text{max}}}^{(k)} - \mathbf{c}_{t_\text{min}}^{(k)}||
\end{equation}

However, this metric alone does not indicate whether an observed topic shift reflects a meaningful directional drift or merely high-dimensional sampling variance. We address this in two ways. First, we assess whether observed topic displacements reflect systematic directional drift by comparing each group’s trajectory in embedding space to a random-walk null. We compute a log–likelihood ratio statistic $\Lambda$ that tests whether the mean step is nonzero given the empirical step covariance: larger $\Lambda$ indicates stronger evidence of persistent, directional movement rather than stochastic fluctuation, with significance quantified by permutation $p$-values. Full methodological details and derivations appear in Appendix~\ref{app:lrt}.  We present significance tests for a selection of cases in our Results.

Second, we conduct a bootstrapped permutation test to ensure the observed drift is a stable phenomenon not introduced by batched, large-scale unsupervised clustering and that resampling recovers the same aligned groups with only small outliers.  We present details and discussion of this secondary test in Appendix~\ref{app:cluster}.

\begin{figure*}[htb]
    \centering
    \includegraphics[width=0.95\linewidth]{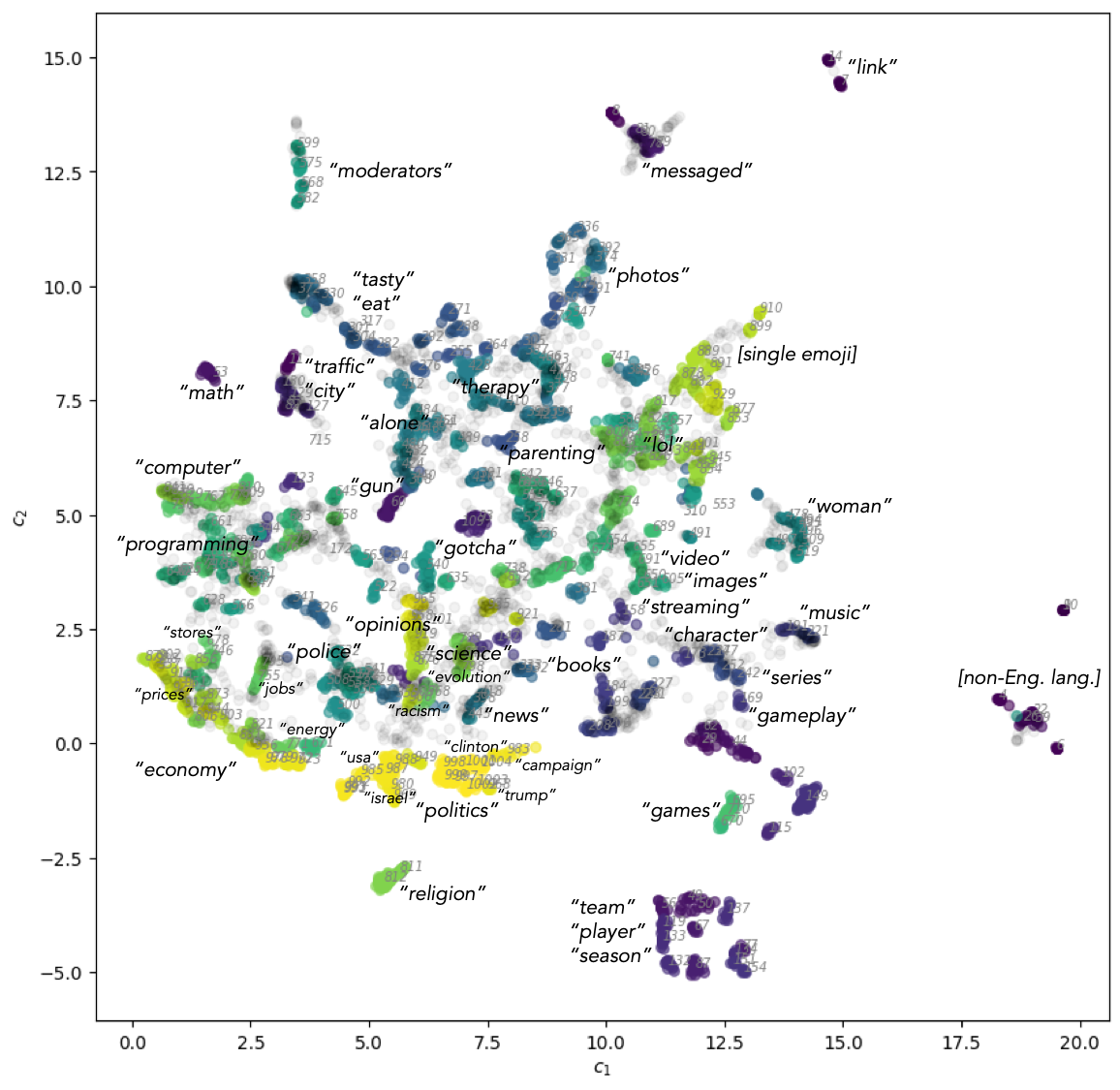}
    \caption{Aligned topic groups over {\it all} time periods of the Reddit corpus (2006-2022) of 12.7 billion comments, using comment-level embeddings, with each dot representing just the topic centroid of a particular time period; labeling is the top shared word across multiple topic groups (using TF-IDF).  Visualization uses a projection to 2-D using UMAP.}
    \label{fig:aligned}
\end{figure*}

\begin{table}[htb]
\centering
\begin{tabular}{ccccc}
 & Topic $k$         & $\delta_k$ & $\Lambda_k$ & $p_k$ \\ \hline
 & ``conspiracy'' & 0.387      & 0.277     & 0.285       \\
 & ``oil''        & 0.374      & 0.794     & 0.295       \\
{\it top} & ``scientific'' & 0.354      & 8.656     & 0.003       \\
 & ``government'' & 0.353      & 2.776     & 0.004       \\
 & ``republican'' & 0.351      & 2.525     & 0.119       \\ \hline
 & ``woman''      & 0.032      & 0.142     & 0.901       \\
 & ``music''      & 0.038      & 0.052     & 0.897       \\
{\it bottom} & ``team''       & 0.044      & 0.113     & 0.942       \\
 & ``games''      & 0.050      & 0.061     & 0.903       \\
 & ``delicious''  & 0.054      & 0.551     & 0.339 
\end{tabular}
\vspace{0.5em}
\caption{Top and bottom-ranked aligned topic groups ordered by {\bf total displacement} ($\delta_k$) in embedding space; log-likelihood ratio ($\Lambda_k$) and computed $p$-value ($p_k$) of nonzero mean drift are also presented.  We see movement in inherently controversial topics like conspiracy theories or politics, while sports or music discussion remains surprisingly stable.}
\label{tab:ranks}
\end{table}

\begin{figure*}[htb]
  \centering
  \begin{minipage}[b]{0.6\textwidth}          
    \centering
    \begin{subfigure}[b]{\linewidth}           
      \includegraphics[width=\linewidth]{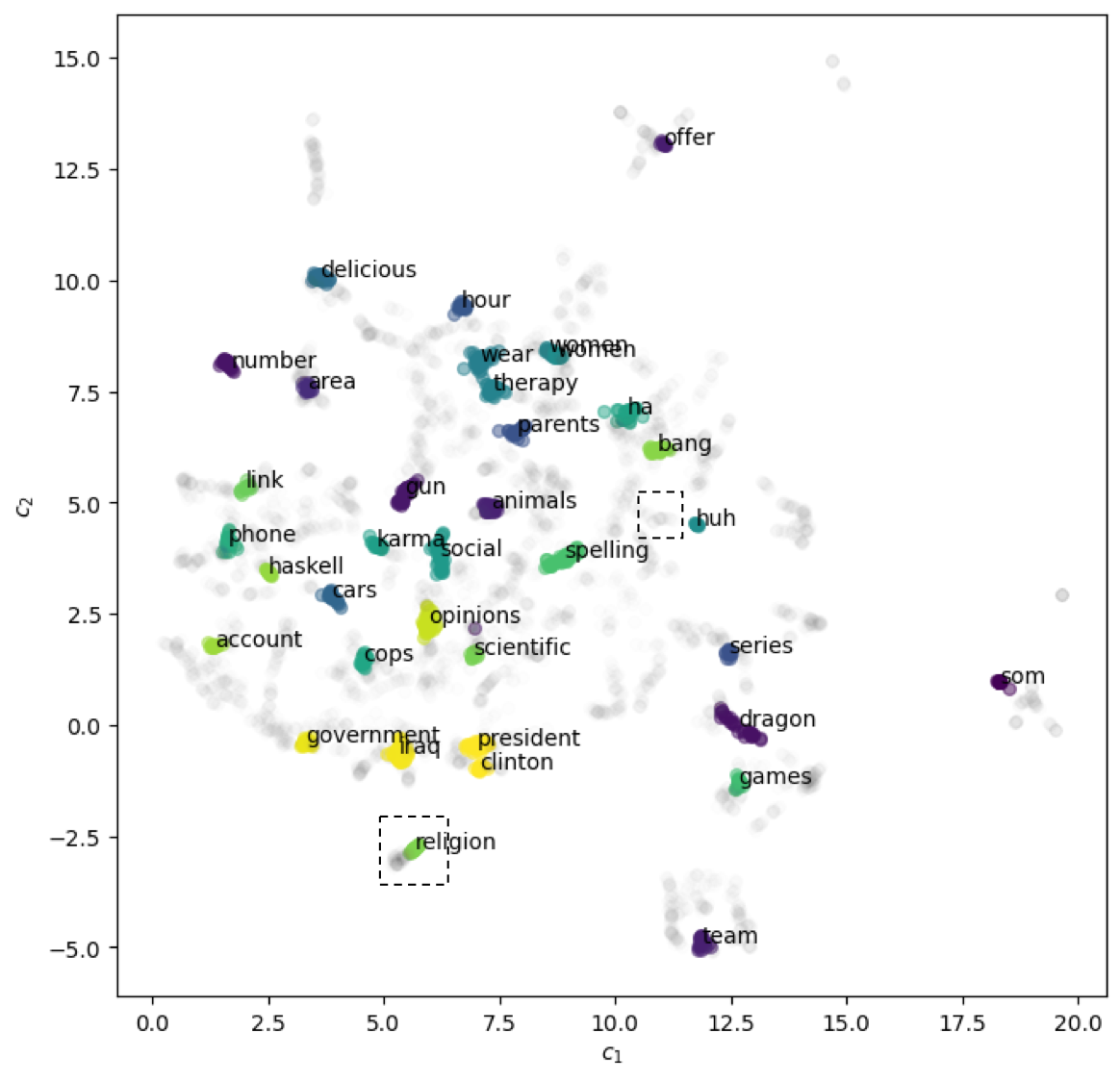}
      \caption{ }
      \label{fig:umap_movers}
    \end{subfigure}
  \end{minipage}%
  \hspace{1em}                                       
  \begin{minipage}[b]{0.28\textwidth}          
    \centering
    \begin{subfigure}[b]{\linewidth}           
      \includegraphics[width=\linewidth]{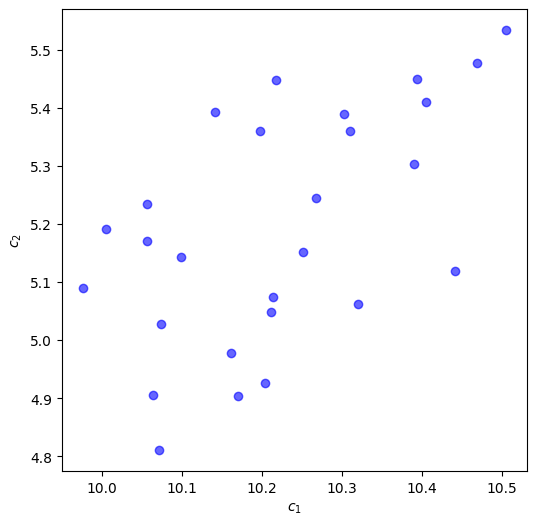}
      \caption{“laugh”}
      \label{fig:zoom_laugh}
    \end{subfigure}\\[1ex]                     
    \begin{subfigure}[b]{\linewidth}
      \includegraphics[width=\linewidth]{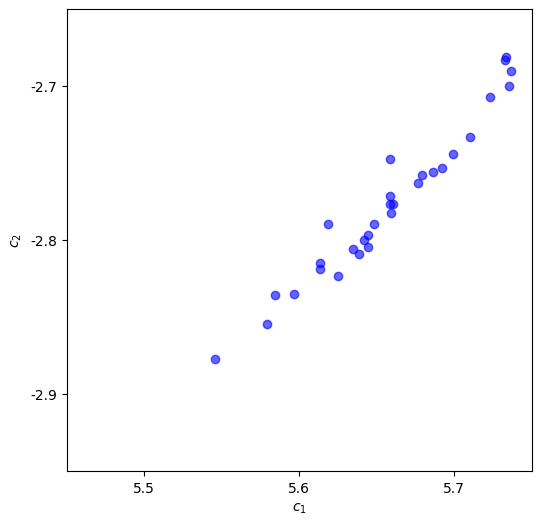}
      \caption{“religion”}
      \label{fig:zoom_religion}
    \end{subfigure}
  \end{minipage}
  \caption{%
    {\bf (a)} Visualization of two regimes of topic groups: those with a significant level of drift (in color) and those without (in gray). 
    {\bf (b)} A topic group related to ``laughing'' exhibiting static movement around a center, with non-significant levels of drift ($\Lambda = 3.74$, $p=0.15$). 
    {\bf (c)} A ``religion'' topic group with significant levels of drift ($\Lambda = 8.29$, $p=0.001$).  All plots are presented in their 2-D UMAP projection.
  }
  \label{fig:rws}
\end{figure*}

\section{Results}\label{results}

\subsection{Topic Detection}

Examining individual monthly windows through TF–IDF-weighted representations, we find that the $k$-means clusters exhibit internal coherence and semantic interpretability, corresponding to major domains of online discourse such as politics, computing, film, dating, travel, and science. Within domains, the method also identifies coherent substructure --- for example, within politics, distinct clusters emerge for electoral events, media organizations, and political figures. This approach also isolates clusters defined by short or single-word responses (e.g., ``ha,'' ``link''). Figure~\ref{fig:singlemonth} illustrates these trends for two example periods, with topics labeled by their top-scoring keywords. Extending the alignment procedure across all time windows (2006–2022) yields $K=1005$ persistent topic groups, the largest extending across 83 monthly clusters. The two-dimensional projections in Figures~\ref{fig:singlemonth} and~\ref{fig:aligned} illustrate that embedding distance corresponds closely to semantic relatedness: topics concerning the arts (music, film, literature) cluster together, as do those on prices and the economy, while more subtle associations --- such as the proximity of discussions on therapy, food, and loneliness --- suggest deeper thematic and contextual linkages.

\begin{figure*}[tb]
\centering
    \begin{subfigure}{0.49\linewidth}
        \centering
        \includegraphics[width=0.85\linewidth]{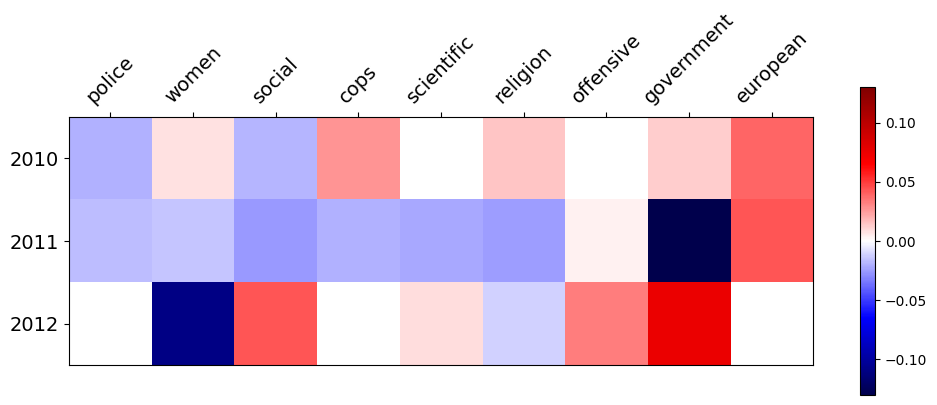}
        \caption{Change relative to ``racism''}
    \end{subfigure} 
    \begin{subfigure}{0.49\linewidth}
        \centering
        \includegraphics[width=\linewidth]{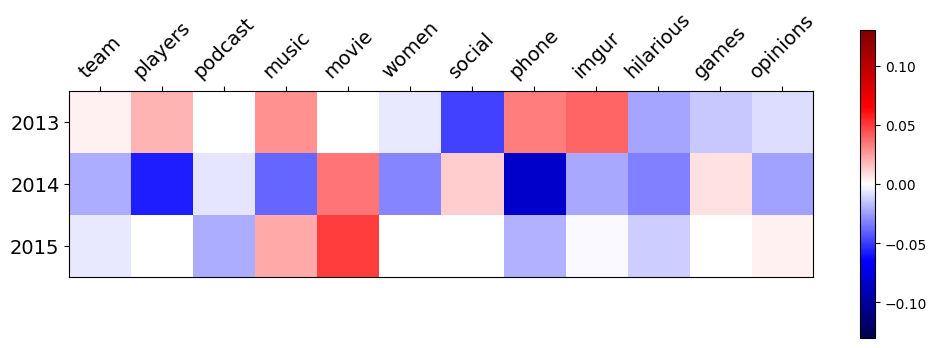}
        \caption{Change relative to ``streaming''}
    \end{subfigure}
    \caption{Average annual change in centroid distance between a selected topic and other nearby topics, annotated by the topic group's top shared keyword across all time periods.  Negative values (blue) mean the topic group is growing closer to the selected topic in embedding space, implying a change in context and semantics.}
    \label{fig:diffs}
\end{figure*}

\subsection{Topic Drift}

The visualization in Fig.~\ref{fig:aligned} indicates that many topics are not static in embedding space over time, suggesting that their underlying semantic content evolves even when clusters remain sufficiently similar to be aligned within the same topic group. Figure~\ref{fig:zoom} illustrates this phenomenon, showing pronounced movement among politically oriented topics compared to the relative stability of a group like ``news.'' To quantify this behavior in the full embedding space ($d=384$), we compute each topic group’s total displacement $\delta_k$, reported in Table~\ref{tab:ranks}. As expected, volatile domains such as politics and conspiracy theory exhibit the largest drift, while topics such as food or music remain nearly stationary.  

Using the random walk framework described in the Methods section, we estimate a likelihood ratio statistic $\Lambda_k$ to test whether each group’s trajectory exhibits systematic (nonzero-mean) drift. Two regimes emerge: topic groups with low $\Lambda_k$ values that fail to reject the null hypothesis of zero-mean movement, and those with significant directional drift ($p<0.05$). The correlation between $\Lambda_k$, $p$-values, and displacement $\delta_k$ highlights that some topics move considerably without coherent directionality, whereas others display persistent, directional evolution in meaning over time (Fig.~\ref{fig:rws}). These results indicate that certain thematic domains undergo significant semantic change (like the groups on government, science, or religion all with large $\Lambda$ and corresponding $p$-value $<0.01$), while others remain more stable (like music or sports or humor).

\subsection{Inter-topic Drift}

We next examine {\it inter}-topic dynamics to determine whether the relative positions of topics in embedding space change systematically over time. Because distance encodes contextual similarity, such movement reflects evolving relationships in discourse --- i.e., if a topic of conversation becomes more or less similar to another topic in its semantics. Figure~\ref{fig:diffs} reports the mean annual change in pairwise distance within the $d=384$ embedding space, restricted to topic pairs within a radius of $0.4$ to ensure semantic relevance.  

\begin{figure}[htb]
    \centering
    \includegraphics[width=\linewidth]{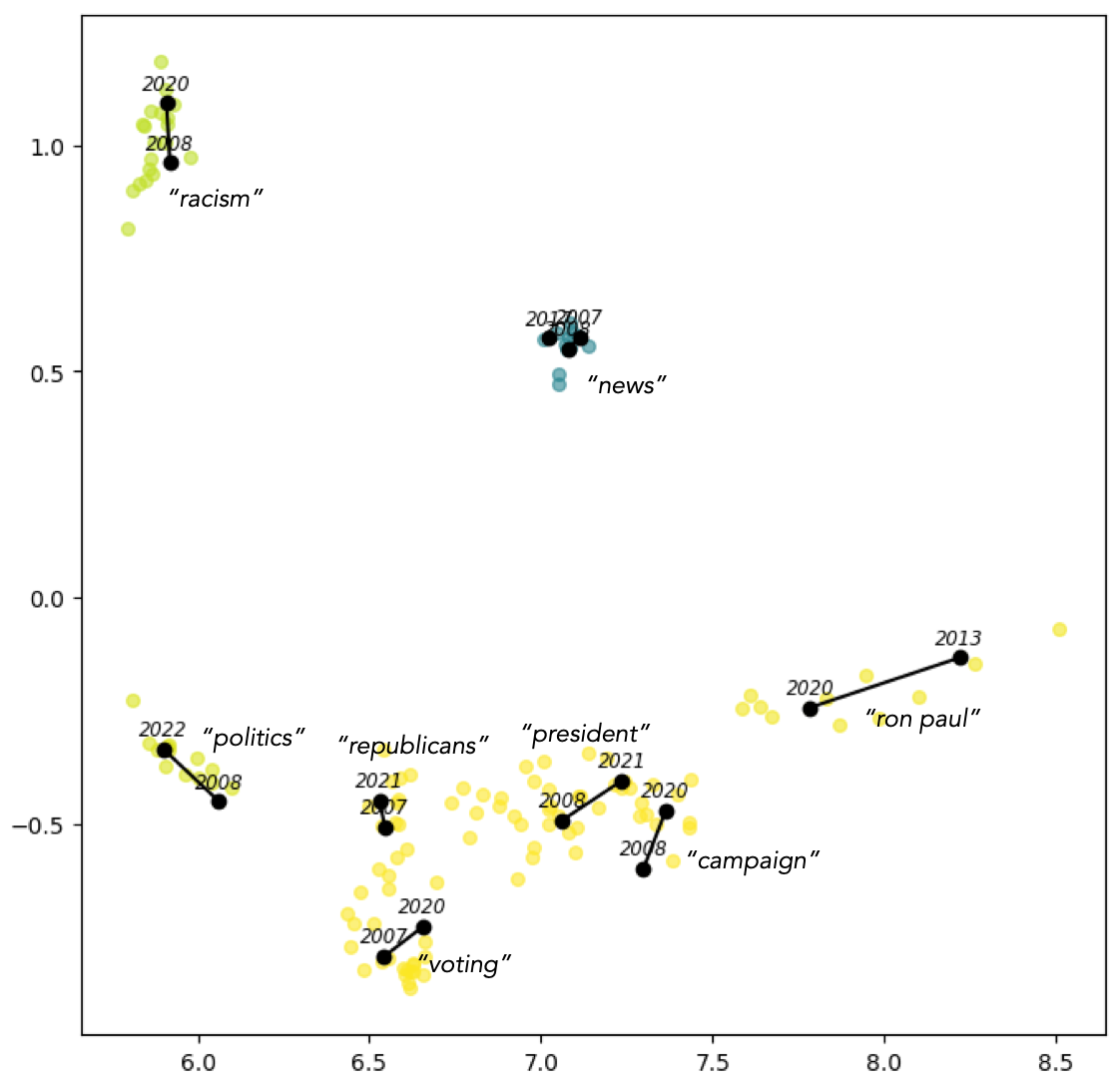}
    \caption{Visualization of several aligned topic groups in a select region of embedding space.  We observe 3-4 topic groups that appear to drift toward a central shared area of embedding space while others move away.}
    \label{fig:zoom}
\end{figure}

As shown in Fig.~\ref{fig:diffs}{\bf (a)}, the topic labeled by ``racism'' has become increasingly proximate to discussions of police, religion, social issues, and women, while distancing from topics on Europeans. A parallel pattern appears in Fig.~\ref{fig:diffs}{\bf (b)}, where ``streaming'' converges toward topics concerning music, video, and mobile technology. Figure~\ref{fig:zoom} further shows several political topics drifting toward an unoccupied region of embedding space, consistent with the emergence of new contextual associations. Together, these results indicate that semantic change occurs not only within topics but also in their relative configuration, reflecting structural evolution in the organization of discourse.

\section{Discussion}

Modern language models enable the representation of language with geometric structure, directly capturing contextual and semantic meaning in an embedded space. This perspective reframes topic modeling as an analysis of movement --- how discourse itself traverses and reshapes semantic space over time. Prior work has explored topic dynamics at the sentence level \cite{grootendorst2022bertopic} and lexical drift at the word level \cite{bamler2017dynamic}; here, we extend both by tracing comment-level trajectories within a large-scale corpus of 12.7\,billion Reddit comments. The results demonstrate that discourse change manifests as measurable geometric motion, both within topics and in their relative configuration. 

This finding carries practical and theoretical significance: computational systems that assume topic stationarity (e.g., retrieval, moderation, classifier calibration) will degrade unless they incorporate temporal geometry, while the inter-topic structure reveals cultural re-association over time. For instance, the convergence of a ``racism'' related topic group toward groups related to ``police,'' ``religion,'' and ``women,'' or of ``streaming'' toward ``music,'' ``video,'' and ``phones,'' reflects durable shifts in meaning rather than transient lexical overlap, offering a quantitative view of how online discourse reconfigures itself.

Three particularly notable patterns emerge from this analysis. First, directional drift is selective, as contentious domains exhibit significant nonzero-mean trajectories (high $\Lambda_k$, low $p_k$), whereas stable domains show near-zero displacement. Second, relational reorganization occurs over time, as inter-topic distances change systematically, suggesting discourse evolves through a reconfiguration of the semantic network rather than independent topic shifts (Fig.~\ref{fig:diffs}). Third, several political topics drift toward previously unoccupied regions of the space (Fig.~\ref{fig:zoom}), indicating convergence toward new contextual meanings not captured by static taxonomies. Together, these results imply that topic movement in embedding space reflects real semantic evolution --- quantifiable changes in how ideas relate to one another and the emergence of new conceptual alignments over time.

Methodologically, treating topics as trajectories with uncertainty-aware tests advances beyond keyword tracking by furnishing falsifiable claims about semantic change. Substantively, the geometry illuminates when concepts gravitate toward new associations (e.g., policing with racism) and when domains decouple, offering evidence for theories of polarization, agenda setting, and technological diffusion. Practically, the statistics we report (displacement, $\Lambda_k$, neighborhood drift) can serve as early-warning indicators for model and policy interventions: large positive neighborhood drift flags domains where curated lexicons or classifier boundaries will become obsolete, while stable regimes justify longer retraining cycles. These use cases point to forecasting and causal attribution as next steps, including dynamical models of attractors and exogenous shocks that can explain—and predict—the observed reconfiguration of discourse.

\subsection{Limitations \& Future work}

Several limitations of the present analysis suggest directions for future research. First, the decision to include the complete Reddit comment corpus with minimal filtering was motivated by a desire for maximal coverage. This, however, entails the presence of automated or low-quality content, whose prevalence has grown substantially over time and may contribute residual noise to the inferred semantic dynamics. Future work could incorporate classifiers to identify or weight such content, allowing assessment of its influence on topic trajectories. Second, although Reddit's scale and topical heterogeneity make it an ideal testbed for large-scale semantic modeling, its discourse conventions are distinctive. Applying this framework to more domain-specific corpora --- such as scientific communication, news archives, or professional forums --- would clarify the extent to which the observed dynamics generalize across linguistic communities. 

A further avenue for improvement lies in methodological refinement of the analysis pipeline. Future work could employ more robust embedding models, enhanced clustering algorithms, or alignment techniques that explicitly account for temporal dependencies in topic structure. Incorporating multilingual or language-aware embeddings could enable a more faithful representation of global discourse and permit analysis of cross-lingual semantic drift.

Another limitation arises from the use of non-deterministic unsupervised clustering methods, such as our batched $k$-means approach and related algorithms like HDBSCAN \cite{grootendorst2022bertopic,rahimi2024antm}. Because these methods are sensitive to data permutations that can influence topic alignment and downstream trajectory estimates, it was necessary to employ both a null-model comparison for statistical validation and bootstrapping tests (Appendix~\ref{app:cluster}) to assess the robustness of our results.  This is both computationally expensive as well as challenging to implement.

This work emphasizes the interpretability and scalability of transformer-based embeddings rather than the formal generative structure of probabilistic topic models \cite{wu2024survey,churchill2022evolution}. This design enables fine-grained, large-scale analysis of discourse but sacrifices some of the inferential depth available in statistical modeling. Future work may bridge these paradigms by combining embedding-based representations with probabilistic or hybrid models to better capture the mechanisms that drive topic formation and semantic change. A dynamical systems perspective, for instance, might reveal whether certain regions of the embedding space act as attractors or repellers. Moreover, integrating datasets with well-defined thematic boundaries --- such as extremist language --- could provide insights into the forces shaping topic trajectories.

\subsection{Conclusion}

In this work, we presented a novel application of dynamic topic modeling to detect and quantify topic drift at the sentence level within a large-scale corpus. By embedding 12.7 billion Reddit comments (2006–2022) in a contextualized semantic space, applying unsupervised clustering, and aligning clusters over different time periods we were able to explore how different topic areas drifted across semantic space over time. 
Our main contribution is a quantitative framework for measuring these shifts directly in embedding space, enabling the detection of significant directional drift both within and between topics.  

Beyond methodological novelty, the findings reveal that linguistic and cultural change can be observed as measurable geometric motion --- an empirical signature of how public discourse reorganizes itself. Topics in politics, social identity, and religion show strong directional drift, whereas domains such as sports and music remain semantically stable, illustrating that social contestation leaves a traceable imprint in language. This framework thus bridges computational linguistics and social theory, offering a reproducible, data-driven approach to studying semantic evolution, polarization, and cultural realignment at scale. Future work may extend these methods toward predictive modeling, identifying the emergence of new conceptual “attractors” and forecasting the trajectories of public discourse across digital platforms.

\bibliography{bibliography}


\newpage

\begin{appendices}

\section{Cluster selection}
\label{app:k}

\begin{figure*}[htb]
    \centering
    \includegraphics[width=\linewidth]{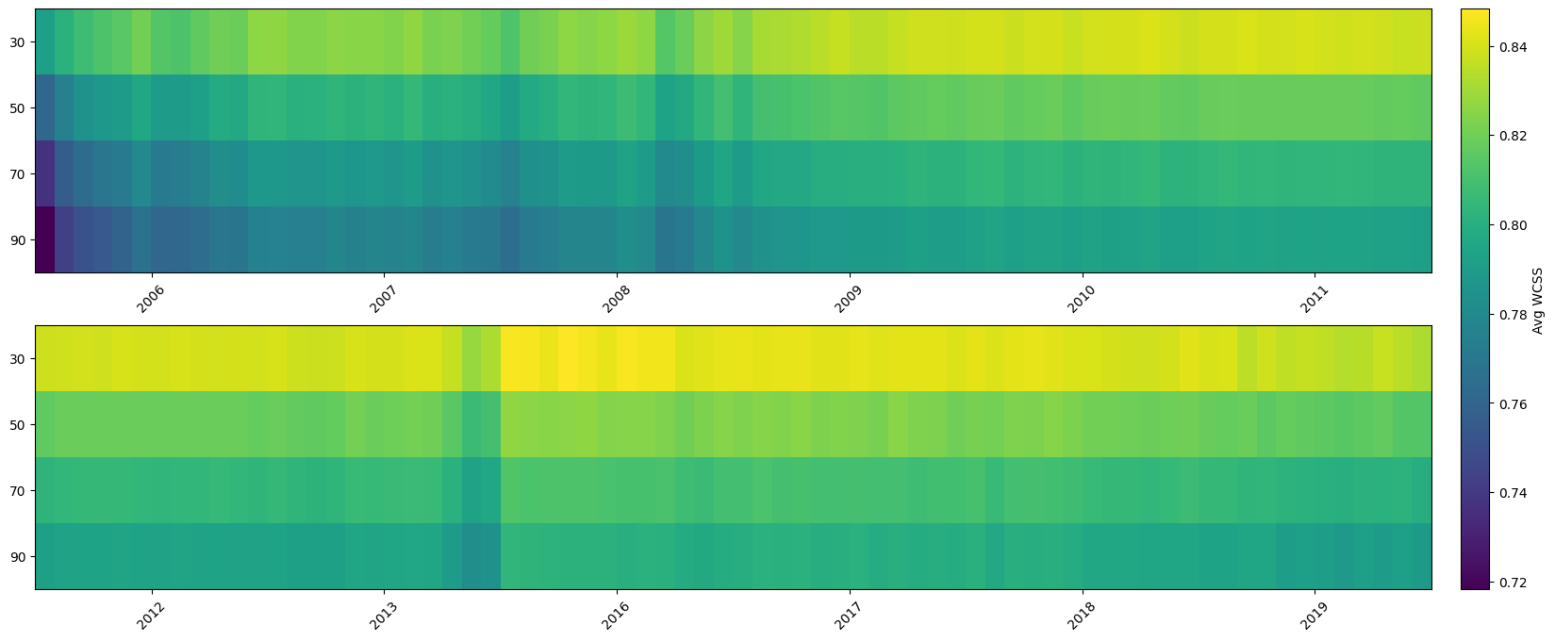}
    \caption{Average within-cluster sum-of-squares (WCSS) for a grid of $k=30,50,70,90$ across all time periods (excluding 2014-2015).}
    \label{fig:kmgrid}
\end{figure*}

\begin{figure}[htb]
    \centering
    \includegraphics[width=\linewidth]{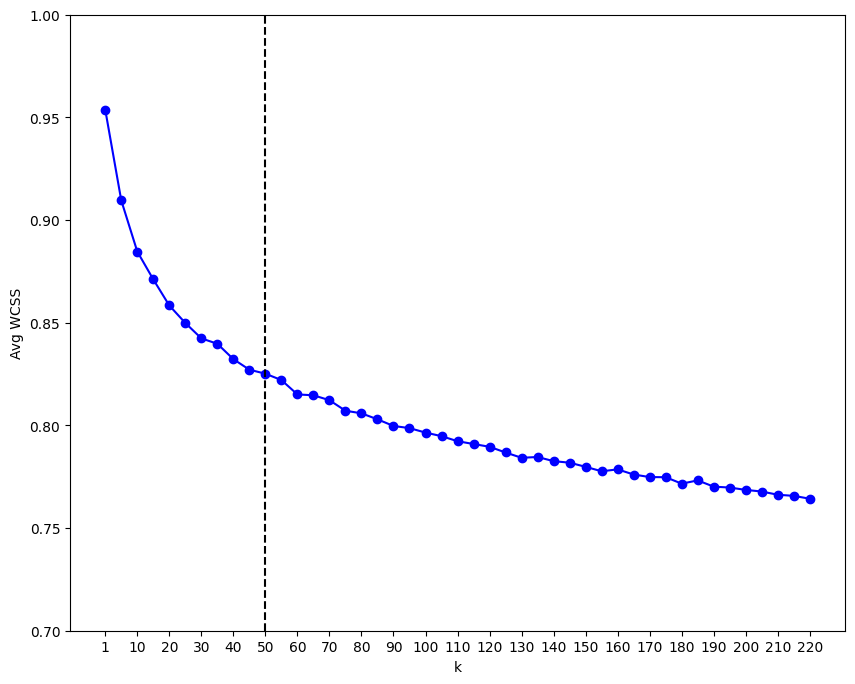}
    \caption{Average within-cluster sum-of-squares (WCSS) for a grid of $1\leq k\leq 220$ on a single demonstration time period (April 2017).}
    \label{fig:kmgrid2}
\end{figure}

Our analysis pipeline, a modification of similar methodology in \cite{reimers2019sentence2,rahimi2024antm}, uses a k-means clustering step following time windowing and before alignment.  We chose to work in the full dimensional space of the sentence embedding in order to avoid prohibitively expensive global dimensionality reduction across the entire dataset, and here present results supporting our choice of $k$.  In the subsequent Appendix we examine the stability of the clusters.

We use the metric of average within-cluster sum-of-squares (WCSS), sometimes termed inertia (e.g. \cite{rykov2024inertia}), as a metric to evaluate appropriate selection of $k$ \cite{ahmed2020k}.  This is simply the average distance between a datapoint and its assigned centroid, or more formally, for a particular time period $t\in\mathcal{T}$:
\begin{equation}
    \text{WCSS} = \frac{1}{N}\sum_{i=1}^N ||\mathbf{e}_i - \mathbf{c}_i||^2
\end{equation}
where we are using $\mathbf{c}_i$ as shorthand for the centroid within $t$ corresponding to the embedding $\mathbf{e}_i$.

We computed this metric for each of the 204 time periods from 2006-2022, across a small grid of $k=30,50,70,90$.  Results are shown in Figure \ref{fig:kmgrid}.  We note remarkable consistency in average inertia across time periods, especially for all time periods 2009 and on.  This indicates a single choice of $k$ is appropriate for all time periods (as opposed to varying $k$ in each time period).

Next, we computed this metric for a single time period across a larger grid of $k$, shown in Figure \ref{fig:kmgrid2}.  We note a quantitatively sharper decrease in $1\leq k\leq 50$, with diminishing returns after $k\approx 50$, leading to a selection of $k=50$ --- this technique is sometimes termed the ``elbow'' criterion \cite{rykov2024inertia}.  Given this and the similarity of results in Figure \ref{fig:kmgrid}, we used $k=50$ for all time periods for the analysis pipeline in the body of the paper.

\section{Cluster validation}
\label{app:cluster}

\begin{figure*}[tb]
\centering
    \begin{subfigure}{0.49\linewidth}
        \centering
        \includegraphics[width=\linewidth]{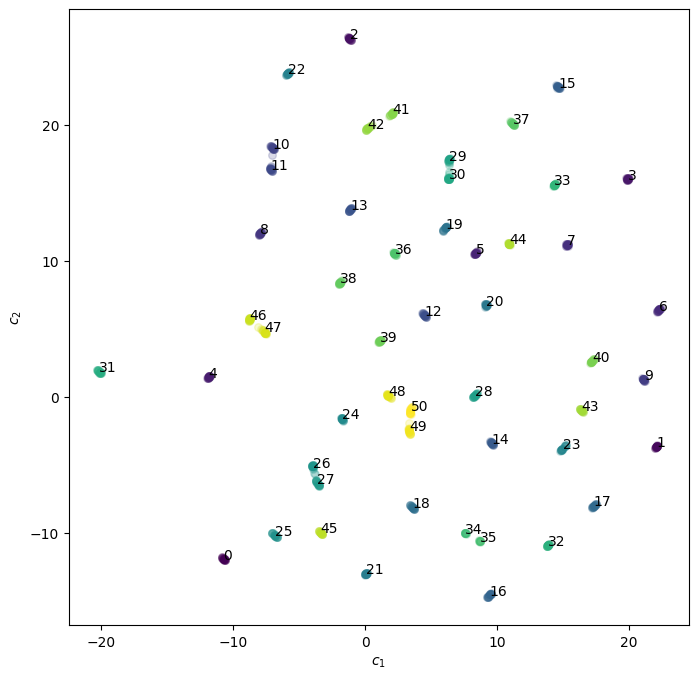}
        \caption{Shared initialization}
    \end{subfigure} 
    \begin{subfigure}{0.49\linewidth}
        \centering
        \includegraphics[width=\linewidth]{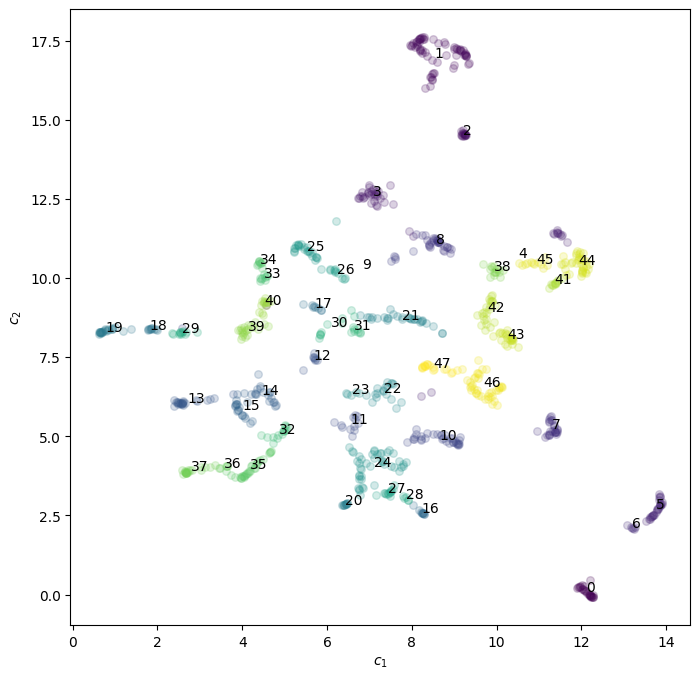}
        \caption{Different initialization}
    \end{subfigure}
    \caption{Aligned topic groups for the same resampled time period, using a single shared centroid initialization in each permutation \textbf{(a)} and different initialization \textbf{(b)}.  Despite variation in centroids between permutations, the method recovers groups that are nearly an exact match for the original clusters, and inspection of the random walk test under both initialization schemes shows no groups have $p$-values below $0.01$.}
    \label{fig:perm}
\end{figure*}

We also explored validation of the stability of the clusters themselves.  Working with such large data necessitates batched and distributed approaches that by nature introduce stochasticity into the otherwise deterministic algorithms underlying k-means \cite{ahmed2020k}.  We seek to validate that the drift within an aligned topic group is not attributable to noisiness introduced in our batch scheme.

To address this, we applied permutation testing to our pipeline.  Specifically, we considered a single time window of data from a central portion of the entire 17-year corpus, resampled it with replacement $R$ times within batches of size $M$, to create $R$ sets of unlabeled embeddings.  We then treat these as $R$ distinct time windows and apply our pipeline: we perform $k$-means clustering on each, then a global dimension reduction on the $R\times k$ learned centroids, then we align similar centroids into $G$ topic groups.  

In a deterministic clustering, with a fixed random seed in initialization, we would expect each permutation to have the exact same centroids, and the resulting aligned groups to be an exact set of $G=k$ groups, each group consisting of $R$ centroids with identical coordinates.  This is the situation in Figure \ref{fig:perm}(a) where we use a single batch over the entire time window and a fixed seed during each initialization (with $R=20$, $k=50$, and a recovered $G=50$).  As further validation, we apply the random walk comparison test outlined in the paper, and find zero groups with $p$-values below 0.05.

Although we are able to follow this deterministic approach for smaller quantities of embeddings, this becomes prohibitive as the size of the time window grows.  In the dataset of this paper, this transition happens around 2015 as the size of each time window grows from hundreds of thousands to tens of millions.  We adopt a batch size of $M=1$ million embeddings and repeat the permutation testing, with the result depicted in Figure \ref{fig:perm}(b).  We still recover $G=49$, though now with some centroids ($\approx 7\%$ of the data) being classified as outliers.  A visual inspection of the 2-dimensional UMAP plot can of course be deceiving, so we again apply the random walk comparison, and find zero groups with $p$-values below 0.01, and only 4 groups with $p$-values below 0.05.

We conclude that although our methodology introduces potential randomness at the clustering step, in large datasets, we here demonstrate is minimized through reasonable batching schemes and filterable by our random walk LRT approach.  We note that the methods proposed in \cite{rahimi2024antm} also introduce potential noise at the dimensionality reduction step prior to clustering, by performing a global dimension reduction using a variant of UMAP called Aligned-UMAP, which we also find prohibitive for datasets at the scale used in this paper.

\section{Random Walk \& LR test}
\label{app:lrt}

This appendix details the test employed to determine the significance of the observed topic displacement over time through comparison to a random walk; see \cite{shumway2000time} for a comprehensive treatment.

Let $\{\mathbf{c}_t\}_{t=1}^{T}$ denote the sequence of topic centroids for a given group, where each $\mathbf{c}_t \in \mathbb{R}^d$ and $T \ge 2$. Define the step (displacement) vectors:
\[
\Delta_t \equiv \mathbf{c}_{t+1}-\mathbf{c}_t \in \mathbb{R}^d,\qquad t=1,\ldots,T-1.
\]
Assume $\Delta_t \stackrel{\text{i.i.d.}}{\sim} \mathcal{N}(\boldsymbol{\mu},\mathbf{\Sigma})$ with unknown mean $\boldsymbol{\mu}\in\mathbb{R}^d$ and positive-definite covariance $\mathbf{\Sigma}\in\mathbb{R}^{d\times d}$. \\
We test:
\[
\begin{aligned}
H_0&:\ \boldsymbol{\mu}=\mathbf{0}\quad (\text{no directional drift}),\\
H_1&:\ \boldsymbol{\mu}\neq \mathbf{0}\quad (\text{systematic drift}).
\end{aligned}
\]

Let $\mathcal{L}(\boldsymbol{\mu}\mid \Delta_{1:T-1})$ denote the likelihood of the observed steps
$\Delta_{1},\ldots,\Delta_{T-1}$ under mean $\boldsymbol{\mu}$ and covariance $\mathbf{\Sigma}$:
\[
\mathcal{L}(\boldsymbol{\mu}\mid \Delta_{1:T-1})
= \prod_{t=1}^{T-1}\mathcal{N}\!\left(\Delta_t;\,\boldsymbol{\mu},\mathbf{\Sigma}\right).
\]

The log–likelihood ratio comparing $H_1$ to $H_0$ simplifies to
\[
\Lambda \;=\; \frac{T-1}{2}\,\bar{\boldsymbol{\mu}}^{\!\top}\mathbf{\Sigma}^{-1}\bar{\boldsymbol{\mu}},
\qquad
\bar{\boldsymbol{\mu}} \equiv \frac{1}{T-1}\sum_{t=1}^{T-1}\Delta_t,
\]
where $\bar{\boldsymbol{\mu}}$ is the empirical mean step, and $\Lambda$ is the log–likelihood ratio statistic comparing $H_1$ to $H_0$. In practice, $\mathbf{\Sigma}$ is replaced by the unbiased sample covariance
\[
\hat{\mathbf{\Sigma}} \;=\; \frac{1}{(T-1)-1}\sum_{t=1}^{T-1}\bigl(\Delta_t-\bar{\boldsymbol{\mu}}\bigr)\bigl(\Delta_t-\bar{\boldsymbol{\mu}}\bigr)^{\!\top}.
\]

To stabilize estimation, the steps are projected via principal component analysis (PCA) onto the first $r=45$ components (explaining $>90\%$ of variance). Denote the projected steps by $\tilde{\Delta}_t\in\mathbb{R}^r$ and the corresponding covariance by $\hat{\mathbf{\Sigma}}_r$; the statistic $\Lambda$ is then computed in $\mathbb{R}^r$ by replacing $(\bar{\boldsymbol{\mu}},\hat{\mathbf{\Sigma}})$ with $(\bar{\boldsymbol{\mu}}_r,\hat{\mathbf{\Sigma}}_r)$.

For each topic group indexed by $k\in\{1,\ldots,K\}$, let $\Lambda_k$ be the observed statistic from its time-ordered steps. A permutation null is obtained by randomly shuffling the time order of $\{\tilde{\Delta}_t\}$ within group $k$ to generate $B=10{,}000$ replicates, yielding $\{\Lambda_k^{(b)}\}_{b=1}^B$. The Monte Carlo $p$-value is
\[
p_k \;=\; \frac{1}{B}\sum_{b=1}^{B}\mathbf{1}\!\left\{\Lambda_k^{(b)} \ge \Lambda_k\right\},
\]
where $\mathbf{1}\{\cdot\}$ denotes the indicator function. Additional diagnostics are reported in Appendix~\ref{app:cluster}.

\end{appendices}





\end{document}